# Real-Time Text Detection and Recognition

Shuonan Pei
*University of Alberta*
*Email: spei@ualberta.ca*

Mingzhi Zhu
*University of Alberta*
*Email: mingzhi4@ualberta.ca*

*Abstract*—**In recent years, Convolutional Neural Network(CNN) is quite a popular topic, as it is a powerful and intelligent technique that can be applied in various fields. The YOLO is a technique that uses the algorithms for real-time text detection tasks. However, issues like, photometric distortion and geometric distortion, could affect the system YOLO accuracy and cause system failure. Therefore, there are improvements that can make the system work better. In this paper, we are going to present our solution - a potential solution of a fast and accurate real-time text direction and recognition system. The paper covers the topic of Real-Time Text detection and recognition in three major areas: 1. video and image preprocess, 2. Text detection, 3. Text recognition. As a mature technique, there are many existing methods that can potentially improve the solution. We will go through some of those existing methods in the literature review session. In this way, we are presenting an industrial strength, high-accuracy, Real-Time Text Detection and recognition tool.**

## 1. Introduction

Real-Time text detection and recognition is a technique of converting the printed or handwritten text into machine readable text in real-time. There are many similar solutions already exist in the market. For example, Real-time Scene Text Detection with Differentiable Binarization[1] or commercial OCR engine like ABBYY FineReader. Object detection is one of the most fundamentation and challenging topics in the text detection field. Text detection faces many problems, such as low resolution, hue, noise, rotation, and etc. Those unpredictable factors fail the outcome result. Therefore, improving the text detection stability and reliability that not only for better user experience , but also for general purposes use within complex environments . In this paper, we will enhance the functionality and the performance of the real-time text detection to perform better by applying approaches such as increasing text detection accuracy with YOLO v4, increasing process speed by pre-processing the input image size, and applying Tesseract latest technology (LSTM) to increase overall text detection accuracy. Additionally, we will conduct literature review to compare existing methods. With the literature review session, our goal is to design a fast and reliable text detection and recognition system. Anyone who has a conventional GPU can use this system. We hope that system can be applied in our daily scenes. Here is our summarized goal:

1) We develop an efficient and powerful text detection model. It allows anyone to use it with a GPU or Neural Network chipset.
2) We verify the existing solutions and techs. By comparing them with our system, any improvement would be acceptable. E.g. accuracy, speed, multi-platform.
3) Compare with existing solutions, list our project advantages and future possible improvements.

## 2. Related Work

### 2.1. Image Pre-processing

In order to remove the noise or enhance the resolution in the image more effectively, many methods based on deep learning are proposed. The currently used approach in this field known as convolutional neural networks. The DnCNN model is able to handle Gaussian denoising with unknown noise level. With the residual learning strategy, DnCNN implicitly removes the latent clean image in the hidden layers [2]. The input of a DnCNN is a noisy image. It focuses on the problem of learning a function F(y) = x to estimate the true clean image. The DnCNN approach adopts the residual learning strategy to train a residual estimate function R(y)= E. The true clean image estimate is then x = y  R(y). The averaged mean squared error between the true residual images and estimates residual from noisy images. The major claims behind DnCNN are the blind denoising capabilities and low time used in the denoising task. The goal of super-resolution (SR) is to recover a high-resolution image from a low-resolution input. The SRCNN is a deep convolutional neural network that learns the end-to-end mapping of low-resolution to high-resolution images [3]. We can use it to improve the image quality of low-resolution images. With a better SR approach, we can get a better quality of a larger image even if we only get a small image originally. The SRCNN consists of the following operations Preprocessing, Feature extraction, Non-linear mapping, and Reconstruction.

#### 2.1.1. Text Detection.

**2.1.1. Text Detection.** The goal of text detection is extracting the required text from images/documents. Often people don't want to read the entire document, rather just a piece of information. Detecting the required text is a tough task but thanks to deep learning, we'll be able to selectively read text from an image. Today text detection can be achieved through two approaches, Region-Based detectors and Single Shot detectors.

In Region-Based methods, the first objective is to find all the regions which have the objects and then pass those regions to a classifier [4]. So it finds the bounding box and afterward, the class of it. This approach is considered more accurate but is slower as compared to the Single Shot approach. Algorithms like Faster R-CNN take this approach. Single Shot detectors predict both the boundary box and the class at the same time. Being a single step process, it is much faster than Region-Based methods [5]. However, Single Shot detectors perform poorly while detecting smaller objects. YOLO is the most popular algorithm of Single Shot detectors. There is a tradeoff between speed and accuracy while choosing the object detector. For example, Faster R-CNN has the highest accuracy, while YOLO is the fastest among all.

### 2.2. Text Recognition

Once we have detected the bounding boxes having the text, the next step is to recognize text. Tesseract is a powerful OCR engine which can be used as a text recognition tool. It was originally developed at Hewlett-Packard Laboratories between 1985 and 1994. Since 2006, it has been maintained by Google. The latest stable version is 4.1.1, released on December 26, 2019 [6]. Tesseract 4.1.1 includes a new neural network, long short term memory(LSTM), based recognition engine that delivers significantly higher accuracy on document images than the previous versions, in return for a significant increase in required compute power. Long short term memory (LSTM) is a special kind of Recurrent Neural Network (RNN) designed to solve the problem of gradient disappearance and explosion during long sequence training[]. Simply say, LSTM performs better in longer sequences than the average RNN[7]. This neural network architecture integrates new feature word and line recognition. The LSTM detects and predicts the text for each frame. The last step is to translate the per-frame predictions into the final label sequence.

## 3. Literature Review

### 3.1. YOLO: An incremental improvement

You only look once (YOLO) is one of the faster object detection algorithms out there. Though it is no longer the most accurate object detection algorithm, it is a very good choice when you need real-time detection, without loss of too much accuracy. YOLO v2 used a custom deep architecture darknet-19, an originally 19-layer network supplemented with 11 more layers for object detection[9]. With a 30-layer architecture, YOLO v2 often struggled with small object detections. YOLO v3 uses a variant of Darknet, which originally has 53 layer network trained on Imagenet[5]. For the task of detection, 53 more layers are stacked onto it, giving us a 106 layer fully convolutional underlying architecture for YOLO v3. This is the reason behind the slowness of YOLO v3 compared to YOLO v2. However, that speed has been traded off for boosts in accuracy in YOLO v3.

The most salient feature of YOLO v3 is that it makes detections at three different scales[5]. YOLO is a fully convolutional network and its eventual output is generated by applying a 1 x 1 kernel on a feature map[8]. In YOLO v3, the detection is done by applying 1 x 1 detection kernels on feature maps of three different sizes at three different places in the network. YOLO v3 makes predictions at three scales, which are precisely given by downsampling the dimensions of the input image by 32, 16, and 8 respectively. Detections at different layers help address the issue of detecting small objects, a frequent complaint with YOLO v2. The upsampled layers concatenated with the previous layers help preserve the fine-grained features which help in detecting small objects. For an input image of the same size, YOLO v3 predicts more bounding boxes than YOLO v2. YOLO v3 predicts boxes at 3 different scales. This means that YOLO v3 predicts 10 times the number of boxes predicted by YOLO v2. You could easily imagine why it's slower than YOLO v2. At each scale, every grid can predict 3 boxes using 3 anchors. Since there are three scales, the number of anchor boxes used in total are 9, 3 for each scale. YOLO v3 performs excellent with other state of art detectors like RetinaNet, while being considerably faster, at COCO mAP 50 benchmark. It is also better than SSD and it's variants.

### 3.2. Efficient and accurate scene text detector

Before the introduction of deep learning in the field of text detection, it was difficult for most text segmentation approaches to perform on challenging scenarios. The conventional approaches are usually multi-staged which ends with slightly lesser overall performance.The EAST algorithm uses a single neural network to predict a word or line-level text[10]. It can detect text in arbitrary orientation with quadrilateral shapes. This algorithm consists of a fully convolutional network with a non-max suppression (NMS) merging state. The fully convolutional network is used to localize text in the image and this NMS stage is basically used to merge many imprecise detected text boxes into a single bounding box for every text region.

The authors have used three branches combining into a single neural network. The first branch is the Feature Extractor Stem. It is used to extract features from different layers of the network. Authors of EAST architecture used PVANet and VGG16 both for the experiment. The second branch is the Feature Merging Branch. It merges the feature outputs from a different layer of the VGG16 network. The input image is passed through the VGG16 model and outputs from different four layers of VGG16 are taken. EAST uses a U-net architecture to merge feature maps gradually. The last branch is the Output Layer. It consists of a score map and a geometry map. The score map tells us the probability of text in that region while the geometry map defines the boundary of the text box. These three branches make the algorithm to be a very robust deep learning method for text detection. EAST can detect text both in images and videos. It runs near real-time at 13FPS on 720p images with high text detection accuracy.

### 3.3. Image Dehazing

The image resolution is influenced by haze which can make targets in images blurred. To resolve this problem, a dehazing method based on the wavelength related physical imaging model is proposed. Colors of objects in an image are decided based on the reflection of different wavelengths, a transmission estimation strategy based on the maximal fuzzy correlation and graph cut result has been designed[11]. The segmentation can be obtained by the maximal fuzzy correlation and the graph cut algorithm, and then it can be used as the guided image in the guided filter for getting a transmission map with continuous scene information. Experimental results show that the haze removal images have higher contrast and better sharpness. The edge and detail features are enhanced significantly. The dehazing effect is remarkable.

### 3.4. Background subtraction

Background subtraction is a technique for separating out foreground elements from the background and is done by generating a foreground mask. There is a new difference clustering method proposed. The problem was framed as clustering the difference clustering vectors between pixels in the current frame and in the background image set. This method analyzes the differences between the current frame and 100 randomly generated frames to get the average difference[12]. To cluster the average difference, 2 clusters were required. They applied the quartile method to remove the potential outliers to get the boundary values. Then, the 2-nearest neighbor algorithm has been applied to the cluster in only one iteration. This not only saves computational time but also achieves high Pr and Fm values for accuracy. This difference clustering method is very practical and competitive for real-time video background subtraction.

Background subtraction problem becomes more challenging once the videos are obtained with a moving camera. To solve the challenging task, Robust Principal Component Analysis (RPCA) is proposed to analyze the motion of video background. RPCA is utilized for subtracting the background of videos obtained from freely moving cameras, in which both angle and magnitude of the optical flow are utilized for the analysis of motion[13]. Finally, superpixels are utilized to compensate for the defects produced by inaccuracies in optical flow. Experimental results show that from standard benchmark datasets this method achieves promising performance.

### 3.5. Active Calibration of Cameras: Theory and Implementation

Text detection and recognition is sensitive with input image quality and bad calibration could cause terrible quality images. Those low quality images impair the final text detection and recognition result. In 1995, A. Basu presented a way to actively calibrate cameras to improve overall image quality[14]. To achieve this, the camera can be mounted on an active platform. Only a scene with strong and stable edges are needed. The outcome is excellent with active calibration camera. We are looking forward to apply this method in our text detection and recognition project to increase the probability of success.

### 3.6. Integrating Active Face Tracking with Model-Based Coding

In reality, objects with text embedded could be rotated and moving. Imagine that, a negligent labour put the licence plate upside down and the driver drove the vehicle without notice. This causes traffic cameras to be unable to read the license plate properly. In paper - Integrating Active Face Tracking with Model-Based Coding, A. Basu and L. Yin presented a topic about a system to track and detect talking faces with a moving and rotated camera[15]. Tracking and recognizing talking faces in video is a challenge due to noise and moving objects. The authors propose step by step to resolve issues. Firstly, background compensation is made to resolve rotation problems. The authors take two images and use a 3D point projected on two image planes with the same lens center. The Algorithms are $x_{t-1} = f\frac{x_t + \alpha \sin\theta y_t + f\alpha\cos\theta}{-\alpha\cos\theta x_t + \gamma y_t + f}$, and $y_{t-1} = f\frac{-\alpha\sin\theta x_t + y_t - f\gamma}{-\alpha\cos\theta x_t + \gamma y_t + f}$. To ensure moving object detection success, A. Basu and L. Yin also presents an advanced head silhouette generation method to track the motion of the talking face. With these concepts and algorithms, our project may be able to identify this rotated moving license plate. However, more investigations and experiments are needed to prove the final outcome.

### 3.7. Hough transform for feature detection in panoramic images

In this paper, the authors discussed Hough transform, a method to recognize geometric shapes from images in image processing[16]. In image processing, it is often necessary to find a straight line that may exist in the scatter set on a plane. That is straight line detection. In text detection and recognition, Tesseract requires line fitting and baseline fitting to accurately locate the line of text. With the Hough transform method, this becomes easy .The easier way to think of this is to select any two scatter points and calculate the k and b coefficient pairs of the linear equation in which they are located. Traversing all scatter two calculations, the higher frequency of the linear equation k, b coefficient pair, that is, the presence of a straight line in the image. However, this method is too violent, when the scatter set is large, the efficiency is very low. The Hough Transform can greatly improve the detection efficiency of straight lines.

### 3.8. Visual gesture recognition for ground air traffic control using the radon transform

Radon Transform is a popular technique which is used in image processing. It can be used in text recognition tasks. In this paper - a complex environment ground air traffic, using Radon Transformation in visual gesture recognition is excellent[17]. Ideally, we can apply Radon transformation to text recognition environments to increase final output.

### 3.9. Gaussian and Laplacian of Gaussian weighting functions for robust feature based tracking

In reality, object tracking could be difficult to achieve due to the complex environment. The author presents Laplacian of Gaussion(LoG) and Gaussian weighting function to improve the Kanade–Lucas–Tomasi (KLT) tracking[18]. By adding two weighting functions to KLT tracking, the tracking result was dramatically improved. Using this theory, we can apply KTL tracking in object recognition and see how it affects the project's result. Future experiments are needed.

### 3.10. A Framework for Adaptive Training and Games in Virtual Reality Rehabilitation Environments

Text detection and recognition not only can be applied to reality, but also can be applied to Virtual Reality(VR). A. Basu presents a low cost way to create a VR world for patient rehabilitation[19]. VR world is highly customizable with proper configuration[19]. With this highly customizable VR world, text tracking and recognition could be added. With this feature enabled, text can be detected and translated into user readable language. Currently, a trained Tesseract supports over hundred languages. We are looking forward to embed our system into the VR world. A low cost and multiple languages supported VR game is achievable.

### 3.11. Eye Tracking and Animation for MPEG-4 Coding

A. Basu proposed an effective way to track eyes' movement[20]. In this paper, Hough transform and deformable templates are applied to track eyes. With the exploitation eyes color information, the speed of tracking is dramatically increased and produces more accurate results. In additional, the author also present a method to track eyes synthesis in the paper. This gave us hint of improving our text tracking in real-time. We need to further investigate into this direction.

## 4. Milestone

1) Test existing solution
2) Rebuild solutions using YOLO and Tesseract with LSTM
3) Improve Accuracy and speed
4) Experiments
5) Report

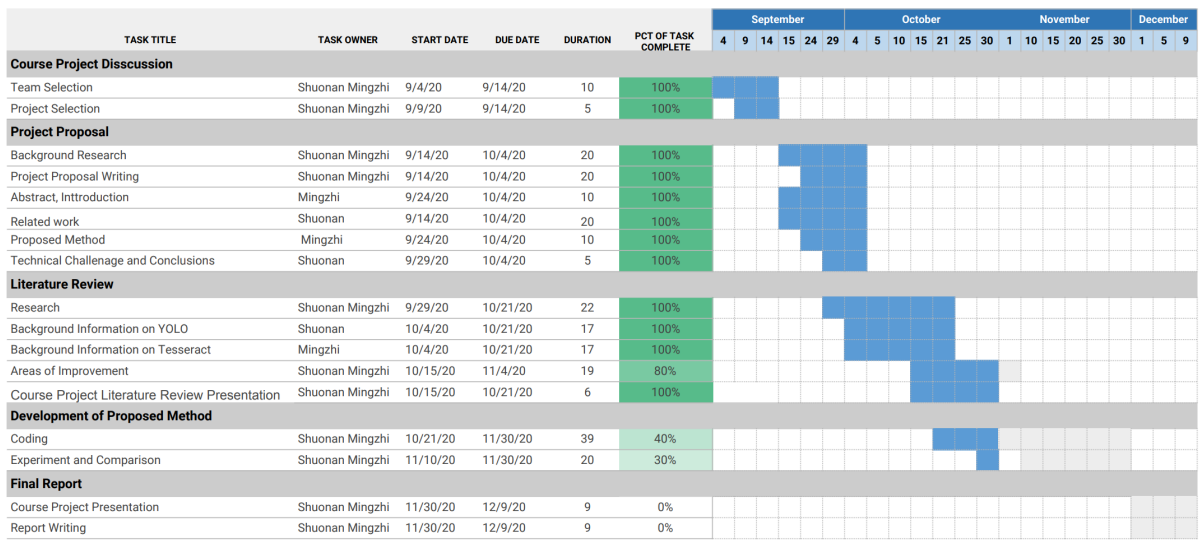

| TASK TITLE | TASK OWNER | START DATE | DUE DATE | DURATION | PCT OF TASK COMPLETE |
|---|---|---|---|---|---|
| **Course Project Disscussion** | | | | | |
| Team Selection | Shuonan Mingzhi | 9/4/20 | 9/14/20 | 10 | 100% |
| Project Selection | Shuonan Mingzhi | 9/9/20 | 9/14/20 | 5 | 100% |
| **Project Proposal** | | | | | |
| Background Research | Shuonan Mingzhi | 9/14/20 | 10/4/20 | 20 | 100% |
| Project Proposal Writing | Shuonan Mingzhi | 9/14/20 | 10/4/20 | 20 | 100% |
| Abstract, Inttroduction | Mingzhi | 9/24/20 | 10/4/20 | 10 | 100% |
| Related work | Shuonan | 9/14/20 | 10/4/20 | 20 | 100% |
| Proposed Method | Mingzhi | 9/24/20 | 10/4/20 | 10 | 100% |
| Technical Challenage and Conclusions | Shuonan | 9/29/20 | 10/4/20 | 5 | 100% |
| **Literature Review** | | | | | |
| Research | Shuonan Mingzhi | 9/29/20 | 10/21/20 | 22 | 100% |
| Background Information on YOLO | Shuonan | 10/4/20 | 10/21/20 | 17 | 100% |
| Background Information on Tesseract | Mingzhi | 10/4/20 | 10/21/20 | 17 | 100% |
| Areas of Improvement | Shuonan Mingzhi | 10/15/20 | 11/4/20 | 19 | 80% |
| Course Project Literature Review Presentation | Shuonan Mingzhi | 10/15/20 | 10/21/20 | 6 | 100% |
| **Development of Proposed Method** | | | | | |
| Coding | Shuonan Mingzhi | 10/21/20 | 11/30/20 | 39 | 40% |
| Experiment and Comparison | Shuonan Mingzhi | 11/10/20 | 11/30/20 | 20 | 30% |
| **Final Report** | | | | | |
| Course Project Presentation | Shuonan Mingzhi | 11/30/20 | 12/9/20 | 9 | 0% |
| Report Writing | Shuonan Mingzhi | 11/30/20 | 12/9/20 | 9 | 0% |

## 5. Conclusion

With YOLO v4 and Tesseract v4, a high speed and accurate real time text detection system can be accomplished. As discussed above, there are many methodologies could possibly apply to our project. The majority challenge is to improve the speed and accuracy of text detection. In this paper, a number of literature are reviewed and discussed. However, industrial strength, high-accuracy, real-time text detection and recognition system is incredibly difficult to achieve. We are greedy to look for better methodologies to improve our work. With more research, I think there are improvements that can be made. We will investigate this direction in our future research to continue to fulfill project expectations and goals.